\documentclass[letterpaper]{article} 
\usepackage{aaai2026}  
\usepackage{times}  
\usepackage{helvet}  
\usepackage{courier}  
\usepackage[hyphens]{url}  
\usepackage{graphicx} 
\usepackage{natbib}  
\usepackage{caption} 
\usepackage{algorithm}
\usepackage{algorithmic}
\usepackage{amsmath}
\usepackage{amssymb}  
\usepackage{multirow}
\usepackage{subcaption}
\usepackage{xcolor}
\usepackage{colortbl}
\usepackage{booktabs} 
\usepackage{adjustbox} 
\usepackage{xcolor} 
\usepackage{subcaption} 
\usepackage{booktabs}   
\usepackage{multirow}   
\usepackage{pifont}     
\usepackage{graphicx}   
\usepackage{amsmath}    

\definecolor{mycolor}{HTML}{FFA500}

\usepackage{newfloat}
\usepackage{listings}
\DeclareCaptionStyle{ruled}{labelfont=normalfont,labelsep=colon,strut=off} 
\floatstyle{ruled}
\newfloat{listing}{tb}{lst}{}
\floatname{listing}{Listing}
\title{ViSP: A PPO-Driven Framework for Sarcasm Generation with Contrastive Learning}

\title{HMID-Net: An Exploration of Masked Image Modeling and Knowledge Distillation in Hyperbolic Space}
\author {
    Changli Wang\textsuperscript{\rm 1},
    Fang Yin\textsuperscript{\rm 2},
    Jiafeng Liu\textsuperscript{\rm 1},
    Rui Wu \textsuperscript{\rm 1, \footnote{Corresponding author: Rui Wu.}},
}
\affiliations {
    \textsuperscript{\rm 1}Faculty of Computing, Harbin Institute of Technology, Harbin, China\\
    \textsuperscript{\rm 2}College of Computer Science and Technology, Harbin University of Science and Technology, Harbin, China
    simple@hit.edu.cn
}

\usepackage{bibentry}

\begin{document}

\maketitle

\begin{abstract}
Visual and semantic concepts are often structured in a hierarchical manner.
For instance, textual concept `cat' entails all images of cats.
A recent study, MERU, successfully adapts multimodal learning techniques from Euclidean space to hyperbolic space, effectively capturing the visual-semantic hierarchy.
However, a critical question remains: how can we more efficiently train a model to capture and leverage this hierarchy?
In this paper, we propose the \textit{Hyperbolic Masked Image and Distillation Network} (HMID-Net), a novel and efficient method that integrates Masked Image Modeling (MIM) and knowledge distillation techniques within hyperbolic space. 
To the best of our knowledge, this is the first approach to leverage MIM and knowledge distillation in hyperbolic space to train highly efficient models. 
In addition, we introduce a distillation loss function specifically designed to facilitate effective knowledge transfer in hyperbolic space.
Our experiments demonstrate that MIM and knowledge distillation techniques in hyperbolic space can achieve the same remarkable success as in Euclidean space.
Extensive evaluations show that our method excels across a wide range of downstream tasks, significantly outperforming existing models like MERU and CLIP in both image classification and retrieval.
\end{abstract}


\section{Introduction}
Humans can perceive the real world through images, where a single image encapsulates a wealth of information. 
This information can be articulated through diverse textual descriptions, each providing a distinct interpretation.
These diverse descriptions exhibit multiple hierarchical relationships.
As humans, we possess the ability to reason from each description and organize the information into coherent visual-semantic hierarchy \cite{ref1,hyper-retrieval1}.

\begin{figure}[tbp]
    \centering
    \begin{minipage}[b]{0.6\linewidth}
        \centering
        \includegraphics[width=\textwidth]{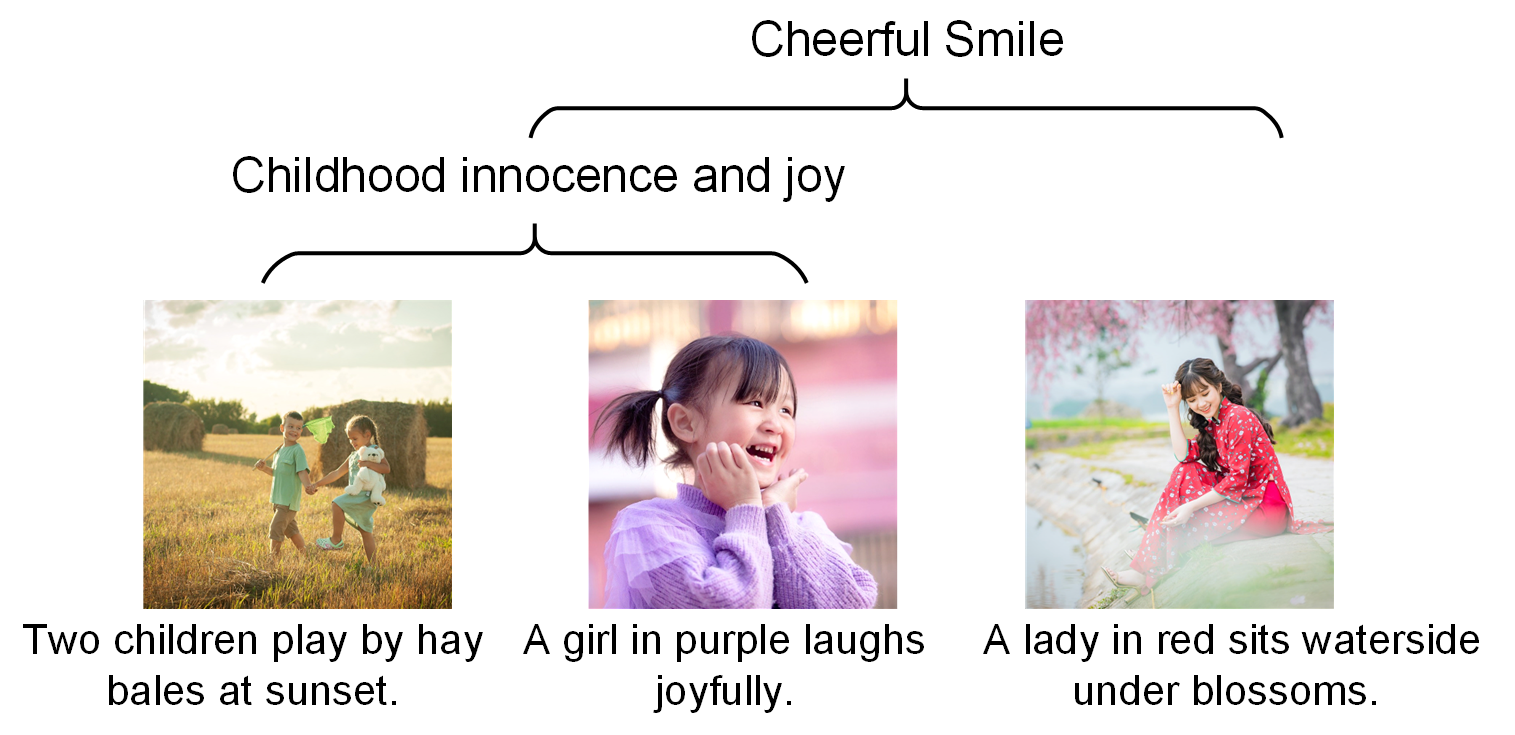}
        {\scriptsize (a) Visual-Semantic hierarchy}
    \end{minipage}
    \begin{minipage}[b]{0.39\linewidth}
        \centering
        \includegraphics[width=\textwidth]{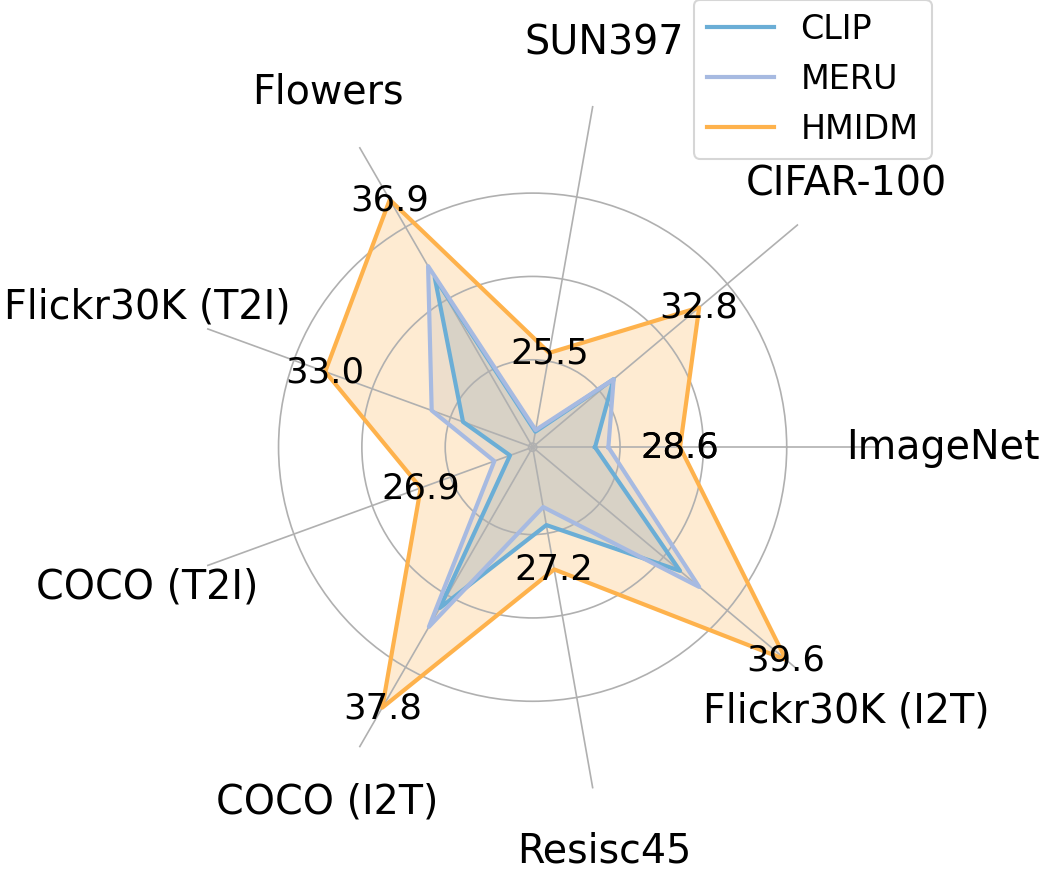}
        {\scriptsize (b) The performance comparison}
    \end{minipage}
    \caption{(a) Images and text descriptions can be viewed as a visual-semantic hierarchy. ``Cheerful smile" is a higher-level concept compared to the image itself, as it can be used to describe smiles of both children and women. (b) presents the performance comparison of HMID-Net, CLIP, and MERU on zero-shot classification and retrieval tasks. HMID-Net significantly outperforms the baselines across various datasets.}
    \label{fig1}
\end{figure}

For instance, the left image in Fig.~\ref{fig1}(a) can be characterized as ``Two children play by hay bales at sunset" or more succinctly as ``Childhood innocence and joy" or ``Cheerful smile".
The visual-semantic hierarchy can be organized as: (Fig.~\ref{fig1}(a) left image) $\rightarrow$ ``Two children play by hay bales at sunset" $\rightarrow$ ``Childhood innocence and joy" $\rightarrow$ ``Cheerful smile".
If multimodal models can effectively capture the hierarchy between vision and semantics, it can further enhance interpretability and generalization.

In recent years, the rapid progression of deep learning has been predominantly fueled by substantial advancements in hardware, enabling the feasibility of large-scale pre-trained Vision-Language Models (VLMs).
Multimodal Large Language Models (MLLMs) have emerged as a central focus of contemporary research.
A range of multimodal models like CLIP \cite{clip} and ALIGN \cite{ALIGN} have emerged and achieved remarkable success on various downstream tasks, such as detection \cite{ref3,ref4}, classification \cite{clip}, and retrieval \cite{ref5, ref6, ref7}. 
In particular, CLIP is trained on a dataset consisting of approximately 400 million image-text pairs, while ALIGN is trained on 1.8 billion image-text pairs, pioneering a new pretraining paradigm and enabling these models to perform a variety of tasks without the need for fine-tuning.
This raises a question:
in the absence of substantial data, how can we effectively train a high-performance model?
Several researchers have adopted techniques such as knowledge distillation \cite{tinyclip,clipkd,textkd}, prompt tuning \cite{coop,cocoop}, and adapter \cite{Tip-Adapter,adapter-retrieval} to mitigate training costs.

However, these methods are all based on Euclidean space, where the capacity of embeddings is linearly tied to their dimensionality, which limits their ability to effectively capture complex data relationships, such as the visual-semantic hierarchy.
In contrast, in hyperbolic space, the capacity increases exponentially with the radius of the sphere, enabling it to accommodate embeddings of any structure while preserving their inherent properties \cite{ref8}.
Currently, hyperbolic space has been widely applied in various fields, including classification \cite{hyper-classfication1,hyper-classfication2,hyper-classfication3,hyper-classfication4}, segmentation \cite{hyper-segmentation}, detection \cite{hyper-detection}, retrieval \cite{hyper-retrieval1,hyper-retrieval2}, and point cloud \cite{hyperbolic-pointcloud}.
\textbf{Similarly, in hyperbolic space, how can we train a more efficient model under conditions of limited data?}
While knowledge distillation have proven effective in Euclidean space, their potential remains underexplored in hyperbolic space.
We identify the following potential reasons:
(1) Most deep learning frameworks and libraries are predominantly designed for Euclidean space, offering limited support for hyperbolic space, significantly increasing the technical difficulty.
(2) Due to the inherent differences in distance measurement between Euclidean and hyperbolic spaces, designing a loss function in hyperbolic space presents unique challenges.
(3) The fundamental operations in hyperbolic space are more intricate and demand significantly greater computational resources.

In this paper, to address these key challenges, we investigate \textbf{Masked Image Modeling} (MIM) and \textbf{knowledge distillation} techniques within hyperbolic space.
To the best of our knowledge, we are the first to apply MIM and knowledge distillation in the hyperbolic space.
This method, called the \textit{Hyperbolic Masked Image and Distillation Network} (HMID-Net), provides a novel approach for investigating the application of MIM and knowledge distillation within hyperbolic space.
Specifically, HMID-Net consists of two components: the student model and the teacher model.
For the input image to the student model, a large proportion of the patches are randomly masked, with only the unmasked patches being fed into the student network, while the entire image is input into the teacher network.
Subsequently, we employ the Exponential map to project the embeddings extracted by both the student and teacher networks into hyperbolic space, obtaining the corresponding hyperbolic embeddings.
In the hyperbolic space, we introduce three loss functions:
(1) \textit{Hyperbolic contrastive learning loss} aligns the image and text embeddings, similar to CLIP.
(2) \textit{Hyperbolic distillation loss} allows the student model to acquire the profound knowledge and reasoning abilities of the teacher model for complex tasks, effectively mitigating performance limitations caused by data scarcity.
(3) \textit{Entailment loss} compels the model to learn the visual-semantic hierarchy, enhancing its ability to perceive and understand the real world.

We validate the effectiveness of HMID-Net on various downstream vision-language (V+L) tasks.
Fig.~\ref{fig1}(b) presents a comparison of HMID-Net with the baseline CLIP and MERU across various benchmarks.
HMID-Net outperforms MERU across 13 out of the 16 datasets for the image classification task, while achieving results comparable to MERU on the remaining two datasets. 
In retrieval tasks, HMID-Net significantly outperforms MERU, achieving +9.9\% improvement on Flickr@10 (I2T), which demonstrates the effectiveness of our approach.

The main contributions of this paper are summarized as follows:

\begin{itemize}
    \item We propose an efficient and straightforward method, called the \textit{Hyperbolic Masked Image and Distillation Network} (HMID-Net). 
    To the best of our knowledge, this is the first implementation of the MIM and knowledge distillation in hyperbolic space.
    \item In hyperbolic space, we propose a knowledge distillation method called Feature Interaction Distillation and derive the associated loss function.
    \item We are also the first to demonstrate the effectiveness of MIM and knowledge distillation in the hyperbolic space, showing that they can achieve the same remarkable success as in the Euclidean space.
    \item  We conduct extensive experiments to thoroughly evaluate the effectiveness of the proposed method, which demonstrates significant improvements and achieves outstanding results across a variety of tasks.
\end{itemize}

\section{Realted Works}
\label{Realted Works} 

\subsection{Masked Image Modeling}
Given the remarkable success of the Masked Language Model (MLM) in Natural Language Processing (NLP), researchers have extensively explored and investigated analogous approaches in vision \cite{mae, simmim, beit, hog}.
MAE \cite{mae} constructs an asymmetric encoder-decoder framework, where the encoder randomly masks and shuffles 75\% of the image, and the decoder is responsible for reconstructing the original pixels.
\cite{ref11} demonstrates through theoretical derivation that MAE can implicitly align masked and unmasked views.
FLIP \cite{flip} applies MAE to multimodal learning and demonstrates that the reconstruction loss and text masking are not necessary.

\subsection{Prompt tuning, Adapter and Knowledge Distillation}
\textbf{Prompt tuning} is a text input segment, such as ``a photo of the large \{\}", that guides a pretrained language model to generate specific outputs or perform tasks. 
It enables task-solving without traditional fine-tuning. 
However, crafting effective manual prompts requires expertise and is highly time-consuming.
CoOp \cite{coop} proposes two learnable prompts: Unified Context and Class-Specific Context, outperforming manual prompts across various domains. 
However, CoOp faces challenges with generalization to unseen categories, a limitation that CoCoOp \cite{cocoop} attributes to overfitting. 

\textbf{Adapter} is a plug-and-play neural network module that requires training only small additional components.
Tip-Adapter \cite{Tip-Adapter} utilizes a query-key caching mechanism, eliminating the need for additional training.
CLIP-Adapter \cite{clip-adapter} proposes integrating an adapter at the end of the backbone network, instead of utilizing prompts, enabling few-shot fine-tuning of the model.
CALIP \cite{calip} enhances CLIP with a parameter-free attention module for cross-modal interaction, eliminating the need for additional downstream data or training.

\textbf{Knowledge Distillation} transfers generalization features from a teacher model to a student model, enabling high performance with reduced computational cost.
\cite{self-distill} introduces self distillation, where the model serves as both the teacher and the student. 
TinyCLIP \cite{tinyclip} introduces affinity imitation and weight inheritance, effectively reducing model size and applying knowledge distillation to CLIP for the first time.
CLIP-KD \cite{clipkd} validates the effectiveness of CLIP knowledge distillation from the perspectives of relationships, features, gradients, and contrastive modes.
However, these methods have not been explored in hyperbolic space.
In this paper, we investigate knowledge distillation within hyperbolic space.

\subsection{Hyperbolic deep neural networks}
In hyperbolic space, there are five well-known isometric models: the Lorentz (Hyperboloid) model, the Poincaré ball model, the Poincaré half-space model, the Klein model, and the Hemisphere model \cite{hyper2}.
\cite{hyper1} uses the Poincaré ball to model hierarchical relationships and applies Riemannian gradient optimization for training.  
\cite{hyper3} reconstructs Euclidean operations (addition, multiplication, FFN) in hyperbolic space.  
\cite{ref8} introduces entailment cones to establish a partial order and express entailment in the Poincaré ball model.
In computer vision, hyperbolic space has a wide range of applications \cite{hyper-classfication1,hyper-classfication2,hyper-classfication3,hyper-classfication4,hyper-detection,hyper-segmentation}.
MERU \cite{hyper-retrieval1} is the first to integrate hyperbolic space into vision-language models (VLMs), with the aim of capturing the visual-semantic hierarchy depicted in Fig.~\ref{fig1}(a).
However, the aforementioned methods cannot directly leverage pre-trained models in Euclidean space. 
In this paper, we employ MIM and knowledge distillation to train an efficient model in hyperbolic space.

\section{Preliminary}
\label{Preliminary}
Hyperbolic geometry is a special case of Riemannian geometry.
Before presenting our method, this section first introduces Riemannian geometry (Section \ref{Riemannian geometry}) and Lorentz model (Section \ref{Lorentz model}).
\subsection{Riemannian geometry}
\label{Riemannian geometry}
\textbf{Manifold}.
A manifold $\mathcal{M}$ of n-dimension is a topological space that, in the neighborhood of each point, is locally approximated by Euclidean space $\mathbb{R}^n$, while globally it may have a more complex structure.

\textbf{Tangent Space.} 
For a point $ p $ on a manifold $ \mathcal{M} $, its tangent space $ T_p\mathcal{M} $ is an $ n $-dimensional vector space that first-order approximates $ \mathcal{M} $ near $ p $.

\textbf{Riemannian Metric.}
For an $n$-dimensional differentiable manifold $\mathcal{M}$, the Riemannian metric $g$ is defined at each point $p \in \mathcal{M}$ as follows:
\begin{equation}
g_p: \mathcal{T}_p\mathcal{M} \times \mathcal{T}_p\mathcal{M} \to \mathbb{R}
\end{equation}
where for any two tangent vectors $v, w \in \mathcal{T}_p\mathcal{M}$, $g_p(v, w)$ provides the angle and length information between them.

\textbf{Riemannian Manifold.}
Riemannian manifold is defined as manifold $ \mathcal{M} $ equipped with Riemannian metric $ g $, which can be represented as the pair $ (\mathcal{M}, g) $.

\textbf{Parallel Transport.}
Parallel transport is a process for transporting tangent vectors along smooth curves, such as geodesics, within a manifold. 
It is formalized as a mapping $ \mathcal{P}_{p \to q} : \mathcal{T}_p\mathcal{M} \to \mathcal{T}_q\mathcal{M} $, which transfers a tangent vector from the tangent space at point $ p $ to the tangent space at point $ q $.

\subsection{Lorentz model}
\label{Lorentz model}
The Lorentz model $ \mathcal{L}^n $ represents $ n $-dimensional hyperbolic geometry, where the hyperbolic space is embedded as a two-sheeted hyperboloid within the $ n+1 $-dimensional Minkowski space.
Formally, it can be expressed as:
\begin{equation}
\mathcal{L}^n = \left \{ \mathbf{x} = (x^0, \ldots, x^n) \in \mathbb{R}^{n+1} : \langle  \mathbf{x},  \mathbf{x} \rangle_{\mathcal{L}} = -1 / c, c > 0 \right\}
\end{equation}
Where $\langle , \rangle_{\mathcal{L}}$ denotes the Lorentz inner product:

\begin{equation}
\langle \mathbf{x}, \mathbf{y} \rangle_{\mathcal{L}} = -x^0 y^0 + \sum_{i=1}^{n} x^i y^i, \quad \mathbf{x} \text{ and } \mathbf{y} \in \mathbb{R}^{n+1}
\label{inner product}
\end{equation}

\textbf{Geodesic.}
A geodesic is the locally shortest path connecting any two points within a space. 
In Euclidean geometry, a geodesic degenerates into a straight line.
The Lorentz distance between two points $x, y \in \mathcal{L}^n$ is defined as:
\begin{equation}
d_{\mathcal{L}}(\mathbf{x}, \mathbf{y}) = \sqrt{1 / c} \cdot \cosh^{-1}(-c \langle \mathbf{x}, \mathbf{y} \rangle_{\mathcal{L}})
\label{geodesic}
\end{equation}

\textbf{Exponential map.}
For $ p \in \mathcal{L}^n $, its tangent space is denoted as $ T_{\mathbf{p}} \mathcal{L}^n $.
The Exponential map provides a way to map the vector $ \textbf{v} $ from the tangent space to the manifold.
The map $ E_{\mathbf{p}} : T_{\mathbf{p}} \mathcal{L}^n \to \mathcal{L}^n $ is defined as:
\begin{equation}
E_{\mathbf{p}}(\mathbf{v}) = \cosh(\sqrt{c} \, \|\mathbf{v}\|_{\mathcal{L}}) \, \mathbf{p} + \frac{\sinh(\sqrt{c} \, \|\mathbf{v}\|_{\mathcal{L}})}{\sqrt{c} \, \|\mathbf{v}\|_{\mathcal{L}}} \, \mathbf{v}
\label{exp}
\end{equation}

\begin{figure*}[tbp]
    \centering
    \begin{minipage}[b]{1.\linewidth}
        \centering
        \includegraphics[width=\textwidth]{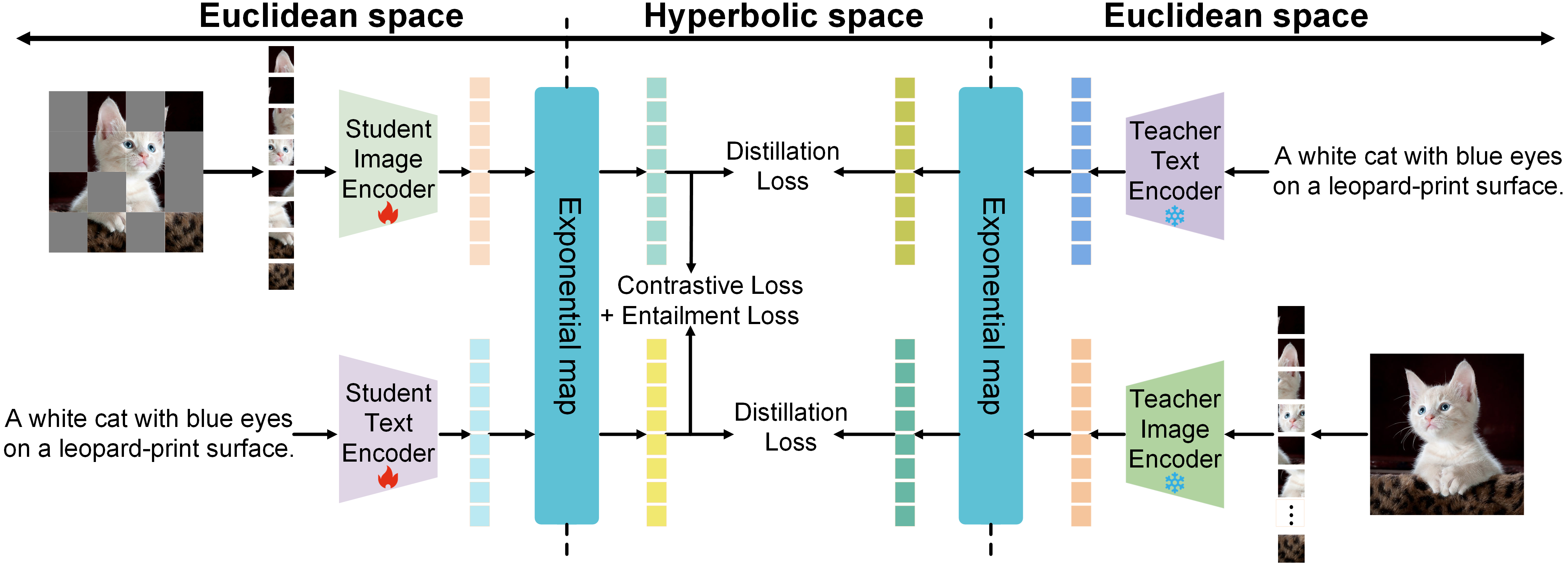}
        {\scriptsize (a) The training process of HMID-Net.}
    \end{minipage}
    \hfill %
    \begin{minipage}[b]{0.97\linewidth}
        \centering
        \includegraphics[width=\textwidth]{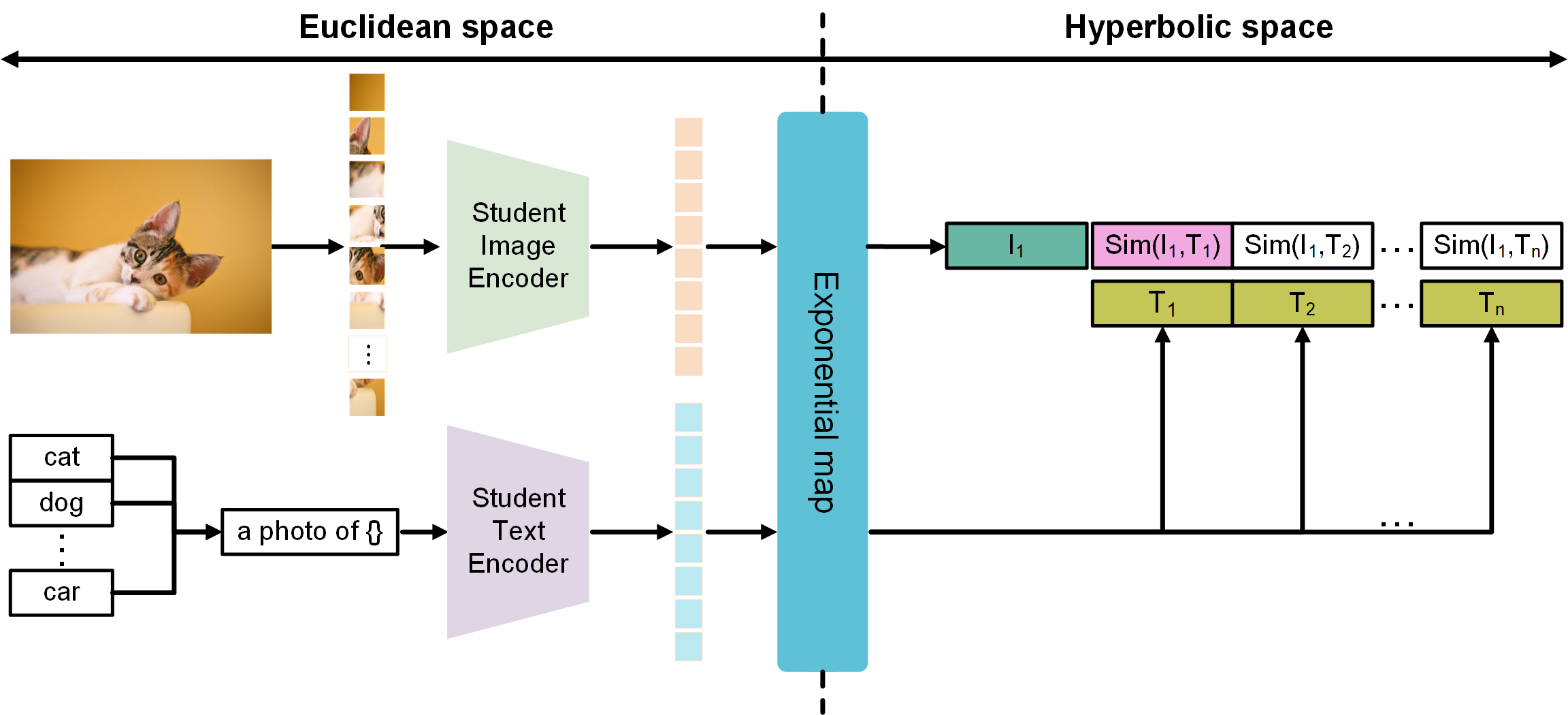}
        {\scriptsize (b) The inference process of HMID-Net.}
    \end{minipage}
    \caption{The overall architecture of HMID-Net. (a) depicts the training process of HMID-Net. In hyperbolic space, the teacher model distills knowledge to the student model, while the contrastive loss aligns the image-text pairs and the entailment loss forces the model to learn the visual-semantic hierarchy. (b) illustrates the inference process of HMID-Net. Unlike CLIP, our inference is performed in hyperbolic space.}
    \label{fig2}
\end{figure*}

\section{Approach}
\label{Approach}

\subsection{A Review of CLIP}
\label{clip}
CLIP \cite{clip} leverages contrastive learning to project images and text into a shared semantic space, allowing the model to effectively capture and understand their semantic relationships.
Specifically, CLIP employs a dual-tower architecture, processing image and text independently through two encoders: the image encoder $h_{\theta}$ and the text encoder $g_{\theta}$.
For an image $ I \in \mathbb{R}^{H \times W \times 3} $, the image encoder utilizes either ResNet \cite{resnet} or ViT \cite{vit} to extract image feature, denoted as $ f(v) = h_{\theta}(I) $.
Similarly, for a text $ T $, the text encoder utilizes a Transformer model to convert the text into text features, denoted as $ f(l) = g_{\theta}(T) $.
During training, image-text pairs within the same batch are positive samples, while unpaired images and text are negative samples.
The model uses a contrastive loss to maximize the alignment between images and their corresponding textual descriptions in the shared semantic space.
The similarity between the image and text embeddings is computed, typically using cosine similarity, and the contrastive loss function $\mathcal{L}_{cl}$ is defined as Eq.~\ref{cl_loss}:

\begin{equation}
sim(f(v),  f(l)) = \frac{f(v) \cdot  f(l)}{\| f(v)\| \|  f(l) \|}
\end{equation}

\begin{equation}
\mathcal{L}_{cl} = -\frac{1}{N} \sum_{i=1}^{N} \log \frac{\exp(\text{sim}(f(v_i),  f(l_i) / \tau)}{\sum_{j=1}^{N} \exp(\text{sim}(f(v_i),  f(l_j) / \tau)}
\label{cl_loss}
\end{equation}

\subsection{Overview of our proposed method}
\label{Overview}
Fig.~\ref{fig2}(a) depicts the training process of our method. 
The architecture consists of a student and a teacher model. 
For a given image-text pair $(I, T)$, the teacher and student models generate embeddings $ [f_\textbf{T}(v),  f_\textbf{T}(l)]$ and $[f_\textbf{S}(v),  f_\textbf{S}(l)]$, respectively.
These embeddings are then projected into hyperbolic space as $[f^{'}_\textbf{T}(v), f^{'}_\textbf{T}(l)]$ and $[f^{'}_\textbf{S}(v), f^{'}_\textbf{S}(l)]$.
In hyperbolic space, contrastive learning is applied between $f^{'}_\textbf{S}(v)$ and $ f^{'}_\textbf{S}(l)$, while the contrastive loss aligns the image-text pairs and the entailment loss forces the model to learn the visual-semantic hierarchy, as detailed in Sections \ref{contrastive learning} and \ref{distillation}.

Fig.~\ref{fig2}(b) depicts the inference process of our method. 
During inference, for a given image-text pair \((I, T)\), the trained student model generates embeddings $ f(v) $ and $ f(l) $ for the image and text, respectively. 
These embeddings are projected into hyperbolic space as $ f^{'}(v) $ and $ f^{'}(l) $.
The similarity between the image and text embeddings is then computed in a manner similar to CLIP.

\subsection{Masked image}
\label{MIM}
We adopt Vision Transformer (ViT) \cite{vit} as the image encoder and the Transformer as the text encoder.
For a given image, it is initially partitioned into non-overlapping patches, with a substantial portion (\textit{e.g.}, 50\%) of the patches randomly masked.
Only the unmasked patches are input into the network, following \cite{mae,flip}.
In this paper, we do not reconstruct the original pixels, as noted in \cite{flip}, because it has minimal impact on the final results and introduces unnecessary computational complexity.

\subsection{Hyperbolic contrastive learning}
\label{contrastive learning}
In hyperbolic space, given a batch of image-text pairs with a batch size of $\mathcal{B}$, the image embedding $ f^{'}(v_i)$ and its corresponding text embedding $ f^{'}(l_i)$ are considered positive samples. 
The remaining $\mathcal{B } - 1$ text embeddings $f^{'}(l_j)$ (where $j \neq i$) in the batch are treated as negative samples.
We adopt the negative Lorentz distance (Eq.~\ref{geodesic}) as the metric for similarity measurement between $ f^{'}(v_i)$ and $ f^{'}(l_i)$.
The logits are scaled by the temperature parameter  $\tau$  to adjust the smoothness of the distribution, after which the softmax function is applied to obtain the normalized probability distribution.
Symmetrically, we also compute the contrastive loss for text.
The contrastive loss $\mathcal{L}_{HCL}$ is computed as the average of the image and text losses for each image-text pair in the batch.


\subsection{Feature Interaction Distillation}
\label{distillation}
To explore effective knowledge distillation methods, \cite{clipkd} first introduced interactive contrastive learning.
In this paper, we extend this method to the hyperbolic space.
Specifically, in the hyperbolic space, given the student image embedding $ f_\textbf{S}^{'}(v) $, student text embedding $ f_\textbf{S}^{'}(l) $, teacher text embedding $ f_\textbf{T}^{'}(l) $, and teacher image embedding $ f_\textbf{T}^{'}(v) $, we replace $ f_\textbf{S}^{'}(l) $ with $ f_\textbf{T}^{'}(l) $ when calculating the image-to-text (I2T) contrastive learning loss, which can be formulated as:
\begin{equation}
\mathcal{L}_{I \to T} = -\frac{1}{N} \sum_{i=1}^{N} \log \frac{\exp(\text{sim}(f_\textbf{S}^{'}(v_i),  f_\textbf{T}^{'}(l_i) / \tau)}{\sum_{j=1}^{N} \exp(\text{sim}(f_\textbf{S}^{'}(v_i),  f_\textbf{T}^{'}(l_j) / \tau)}
\label{I to T}
\end{equation}

Similarly, when calculating the text-to-image (T2I) contrastive learning loss, $ f_\textbf{T}^{'}(v) $ is used to replace $f_\textbf{S}^{'}(v) $, which can be formulated as:
\begin{equation}
\mathcal{L}_{T \to I} = -\frac{1}{N} \sum_{i=1}^{N} \log \frac{\exp(\text{sim}(f_\textbf{S}^{'}(l_i) , f_\textbf{T}^{'}(v_i) / \tau)}{\sum_{j=1}^{N} \exp(\text{sim}( f_\textbf{S}^{'}(l_i) , f_\textbf{T}^{'}(v_j) / \tau)}
\label{T to I}
\end{equation}

The Feature Interaction Distillation loss can be formulated as:
\begin{equation}
\mathcal{L}_{DL} = \frac{1}{2} (\mathcal{L}_{I \to T} + \mathcal{L}_{T \to I})
\label{T to I}
\end{equation}

\begin{figure}[tbp]
\centering
\includegraphics[width=0.4\textwidth]{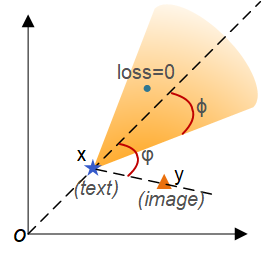}
\caption{Entailment loss. This loss pushes \( y \) into the cone formed by \( x \) to satisfy the partial order. If \( y \) resides within the cone, the loss is equal to zero.}
\label{entailment}
\end{figure}

\begin{table*}[tbp]
    \centering
     \scriptsize
    \begin{adjustbox}{max width=\textwidth}
        \begin{tabular*}{\textwidth}{@{\extracolsep{\fill}}lllll}
            \toprule
            Dataset & Classes & Train & Val & Task \\ 
            \midrule
            ImageNet \cite{imagenet} & 1000 & 1,281,167 & 50,000 & General object classification \\
            Food101 \cite{food} & 101 & 75,750 & 25,250 & Fine-grained classification \\
            CIFAR10 \cite{cifar} & 10 & 50,000 & 10,000 & General object classification \\
            CIFAR100 \cite{cifar} & 100 & 50,000 & 10,000 & General object classification \\
            SUN397 \cite{sun} & 397 & 76,128 & 19,849  & Scene recognition \\
            Aircraft \cite{aircraft} & 100 & 3,334 & 3,333 & Fine-grained classification \\
            DTD \cite{dtd} & 47 & 1,880 & 1,880 & Fine-grained classification \\
            Pets \cite{pets} & 37 & 3,680 & 3,669 & Fine-grained classification \\
            Caltech101 \cite{Caltech101} & 102 & 3,060 & 6,084 & General object classification \\
            Flowers \cite{Flowers} & 102 & 1,020 & 6,149 & Fine-grained classification \\
            STL10 \cite{stl10} & 10 & 5,000 & 8,000 & General object classification \\
            Resisc45 \cite{Resisc45} & 45 & \rule{0.35cm}{0.4pt} &  25,200 & Remote sensing classification \\
            Country211 \cite{clip} & 211 & \rule{0.35cm}{0.4pt} & 21,100  & General object classification \\
            MNIST \cite{minist} & 10 & 60,000 & 10,000 & Handwritten digit classification \\
            CLEVR \cite{clevr} & 8 & \rule{0.35cm}{0.4pt}  & 5,000 & Visual question answering \\
            SST2 \cite{clip} & 2 & \rule{0.35cm}{0.4pt} & 1,821 & Sentiment analysis \\
            \midrule
            COCO \cite{coco} & \rule{0.35cm}{0.4pt} & 118,000 & 5,000 & Image and text retrieval \\
            Flickr30K \cite{Flickr30k} & \rule{0.35cm}{0.4pt} & 29,000 & 1,000 & Image and text retrieval \\
            \bottomrule
        \end{tabular*}
    \end{adjustbox}
    \caption{Details of the dataset utilized in the experiment. The first 16 datasets are employed for zero-shot image classification, while the latter two are utilized for zero-shot image and text retrieval.}
    \label{datasets}
\end{table*}

\subsection{Entailment loss}
\label{Entailment loss}
\cite{ref1} proposes utilizing partial orders to represent the relationship between vision and text.
\cite{ref8} introduces the Entailment loss to learn image-text pairs and their partial order relationships. 
Similar to \cite{hyper-retrieval1}, we incorporate the entailment loss to enhance the model's ability to capture visual-semantic hierarchy.

Fig.~\ref{entailment} illustrates the principle of the Entailment loss.
Let $\mathbf{x} = [x^0, \tilde{\mathbf{x}}]$ and $\mathbf{y} = [y^0, \tilde{\mathbf{y}}]$, where $\mathbf{x}, \mathbf{y} \in \mathcal{L}^n$, and $\tilde{\mathbf{x}} = (x^1, \dots, x^n)$, $\tilde{\mathbf{y}} = (y^1, \dots, y^n)$.
For each \( \mathbf{x} \), the half-aperture of the cone is defined as in Eq. \ref{half-aperture}. 
For the exterior angle between $ \mathbf{x}$ and $ \mathbf{y} $, $ \varphi(\mathbf{x}, \mathbf{y}) = \pi - \angle O\mathbf{x}\mathbf{y} $, as defined in Eq. \ref{exterior}.

\begin{equation}
\phi(\mathbf{x}) = \sin^{-1} \left( \frac{2K}{\sqrt{c} \, \|\tilde{\mathbf{x}}\|} \right)
\label{half-aperture}
\end{equation}

\begin{equation}
\varphi(\mathbf{x}, \mathbf{y}) = \cos^{-1} \left( \frac{y_0 + x_0 \, c \, \langle \mathbf{x}, \mathbf{y} \rangle_{\mathcal{L}}}{\|\tilde{\mathbf{x}}\| \sqrt{(c \, \langle \mathbf{x}, \mathbf{y} \rangle_{\mathcal{L}})^2 - 1}} \right)
\label{exterior}
\end{equation}
Where $ c $ represents the curvature, and $ K $ is a constant with a value of 0.1.

When the exterior angle $\varphi$ is smaller than half the aperture of the cone $\phi$, it indicates that $\textbf{x}$ and $\textbf{y}$ satisfy the partial order relation, in which case no penalty is imposed.
However, if the exterior angle $\varphi$ exceeds the half-aperture of the cone $\phi$, a penalty is imposed to enforce the partial order constraint. 
The Entailment loss is defined as follows:
\begin{equation}
\mathcal{L}_{EL} = \max(0, \, \varphi(\mathbf{x}, \mathbf{y}) - \phi(\mathbf{x}))
\label{el}
\end{equation}

\subsection{Overall Loss}
\label{loss}
We combine all the loss functions to obtain the final loss function $\mathcal{L}$, as shown in Eq.\ref{loss}, enabling the joint training of the model.
\begin{equation}
\mathcal{L} = \mathcal{L}_{HCL} +\lambda_{distillation}\mathcal{L}_{DL}+\lambda_{entailment}\mathcal{L}_{EL}
\label{loss}
\end{equation}

\begin{table*}[tbp]
    
    \centering
    \setlength{\arrayrulewidth}{6mm} 
    \begin{center} 
    \resizebox{\textwidth}{!}{
    
        \Large 
        \begin{tabular}{lcccccccccccccccccc}
            \toprule
            &  & \rotatebox{90}{ImageNet} &
            \rotatebox{90}{Food101} &
            \rotatebox{90}{CIFAR10} &
            \rotatebox{90}{CIFAR100} &
            \rotatebox{90}{SUN397} &
            \rotatebox{90}{Aircraft} &
            \rotatebox{90}{DTD} &
            \rotatebox{90}{Pets} &
            \rotatebox{90}{Caltech101} &
            \rotatebox{90}{Flowers} &
            \rotatebox{90}{STL10} &
            \rotatebox{90}{Resisc45} &
            \rotatebox{90}{Country211} &
            \rotatebox{90}{MNIST} &
            \rotatebox{90}{CLEVR} &
            \rotatebox{90}{SST2} \\ 
            \midrule
            \multirow{3}{*}{ViT S/16} 
            & CLIP & 20.2 & 54.5 & \textbf{49.7} & 18.9 & 18.3 & \textbf{1.3 }& 10.6 & 52.6 & 44.4 & 32.2 & 81.6 & 21.4 & 3.3 & 10.0 & 11.5 & 51.5 \\
            & MERU  & 21.0 & 57.4 & 48.8 & 17.1 & 18.5 & 0.9 & 9.1 & 50.2 & 43.7 & 31.2 & 81.6 & 20.0 & 3.2 & 12.6 & 12.4 & 50.3 \\
            & HMID-Net  & \textbf{25.6} & \textbf{61.6} & 45.3 & \textbf{19.6} & \textbf{22.2} & 1.2 & \textbf{14.6} & \textbf{60.6} & \textbf{51.6} & \textbf{38.1} & \textbf{83.2} & \textbf{22.8} & \textbf{3.7} & \cellcolor{mycolor}\textbf{12.8} & \textbf{13.6} & \textbf{52.1}  \\
            \midrule
            \multirow{3}{*}{ViT B/16} 
            & CLIP & 23.1 & 65.1 & 54.0 & 24.5 & 21.5 & \textbf{1.4} & 11.6 & 59.4 & 52.2 & 38.9 & 82.7 & 21.8 & 3.6 & 9.2 & 13.8 & 51.0 \\
            & MERU  & 22.5 & 61.6 &\textbf{ 55.2 }& 18.0 & 19.2 & 1.5 & 10.2 & 58.7 & 45.5 & 34.6 & 83.0 & 20.0 & 3.3 & \textbf{9.5} & 10.9 & 50.5\\
            & HMID-Net  & \textbf{27.7} & \cellcolor{mycolor}\textbf{66.9} & 49.1 & \textbf{28.5} & \textbf{24.8} & \textbf{1.4} & \cellcolor{mycolor}\textbf{16.8} & \cellcolor{mycolor}\textbf{65.2} & \textbf{53.6} & \cellcolor{mycolor}\textbf{42.2} & \textbf{85.1} & \textbf{25.2} & \textbf{4.0} & 9.1 & \cellcolor{mycolor}\textbf{20.8} & \cellcolor{mycolor}\textbf{52.4} \\
            \midrule
            \multirow{3}{*}{ViT L/16} 
            & CLIP & 23.5 & 60.6 & \cellcolor{mycolor}\textbf{62.2} & 26.1 & 20.7 & 0.7 & 10.2 & 60.2 & 51.8 & 31.5 & 85.8 & 24.5 & 3.4 & \textbf{10.1} & 11.2 & \textbf{50.7} \\
            & MERU  & 24.3 & 63.1 & 61.4 & 26.1 & 20.8 & 1.3 & 11.6 & 62.0 & 52.9 & 32.3 & 85.5 & 23.4 & 3.8 & 9.6 & 12.5 & 50.0 \\
            & HMID-Net   & \cellcolor{mycolor}\textbf{28.6} & \textbf{66.8} & 59.1 & \cellcolor{mycolor}\textbf{32.8} &\cellcolor{mycolor}\textbf{ 25.5} & \cellcolor{mycolor}\textbf{1.8} & \textbf{14.6} & \textbf{ 63.5} & \cellcolor{mycolor}\textbf{59.4} & \textbf{36.9} & \cellcolor{mycolor}\textbf{89.3} & \cellcolor{mycolor}\textbf{27.2} & \cellcolor{mycolor}\textbf{4.5} & 9.4 & \textbf{13.9} & 50.0 \\
            \bottomrule
        \end{tabular}    }
    \caption{\textbf{Zero-shot image classification.} HMID-Net significantly outperforms the baseline CLIP and MERU on 13 out of the 16 datasets. The best performance in each column is highlighted with \textcolor{mycolor}{orange}.}
    \label{zero-classification}
    \end{center} 
\end{table*}
\section{Experiments}
\label{Experimental results}
In this section, we assess the performance of our method across diverse downstream vision-language (V+L) tasks.
Additionally, in Section \ref{ablations}, we conduct comprehensive ablation studies to assess the impact of each component on the overall performance.

\subsection{Implementation details}
\label{Implementation details}
\textbf{Baselines.} 
We first compare our method with CLIP \cite{clip}, which effectively achieves joint multimodal representation by embedding images and text into Euclidean space.
The primary focus of our work is a comparison with MERU \cite{hyper-retrieval1}, which explores the representation of images and text in the hyperbolic space for the first time.
Similar to MERU, we pretrain our model using the Redcaps dataset, which contains $\sim$12 million image-text pairs collected from 350 manually selected subreddits on Reddit.
Due to the absence of some data on the website,  we are only able to download $\sim$ 7 million image-text pairs. 
Consequently, we retrain MERU and CLIP using the available dataset.

\textbf{Datasets.} 
To evaluate our method, we select 18 datasets, including 16 for image classification and 2 for image-text retrieval.
The image classification tasks span a wide range of domains, covering general object classification (ImageNet \cite{imagenet}, Caltech101 \cite{Caltech101}), fine-grained object classification (Food101 \cite{food}, Pets \cite{pets}, Flowers \cite{Flowers}), scene recognition (Sun397 \cite{sun}), and remote sensing classification (Resisc45 \cite{Resisc45}), among others.  
For image-text retrieval tasks, we utilize the COCO \cite{coco} and Flickr30K \cite{Flickr30k} datasets.
A comprehensive overview of these datasets is provided in Table \ref{datasets}.

\textbf{Models.} 
We adopt ViT \cite{vit} as the image encoder and select three different variants: ViT-S, ViT-B, and ViT-L.
The patch size for all models is set to 16. 
For the text encoder, we use a 12-layer, 512-dimensional Transformer \cite{transformer}, consistent with the implementation of MERU.

\textbf{ Initialization.}
We adopt the same initialization strategy as MERU, where the position embeddings remain frozen during training. 
The temperature parameter in the contrastive loss is initialized as $\tau = 0.7$, with a minimum value set to $\tau_{\text{min}} = 0.01 $.
Additionally, the curvature $ c $ of the hyperbolic space is treated as a learnable parameter, initialized to $ c = 1.0 $, with an upper bound of $ c_{\text{max}} = 10 $ to maintain training stability.
The hyperparameters are configured as $ \lambda_{\text{distillation}} = 1 $ and $ \lambda_{\text{entailment}} = 0.2 $ in Eq.~\ref{loss}.

\textbf{Training details.}
We utilize the publicly available Open-CLIP \cite{open-clip} as the teacher model.
During training, the image and text encoders of the teacher model are frozen, while the parameters of the image and text encoders of the student model are updated.
The embeddings are projected into the hyperbolic space using the Exponential map, with both the teacher and student models sharing the same curvature $ c $.
We use the AdamW \cite{adaw} optimizer to train the model, with a weight decay of 0.2 and a maximum learning rate of $5 \times 10^{-4} $.
The learning rate undergoes linear growth during the first 10\% of the total iterations, followed by a cosine decay until it reaches zero.
All models in this paper are trained for 560,000 iterations (approximately 20 epochs) with a batch size of 256.
The implementation is based on PyTorch, and the training is conducted on four NVIDIA GeForce RTX 4090 GPUs.

\subsection{Image classification}
\label{classification}
In image classification, CLIP-style methods utilize prompts to convert predefined labels into textual embeddings for processing by the text encoder.
Subsequently, the similarity between the image and text embeddings is computed, with the textual embeddings exhibiting the highest similarity designated as the predicted outcome.

\begin{figure*}[tbp]
    \begin{minipage}[b]{0.5\linewidth}
        \centering 
        \includegraphics[width=\textwidth]{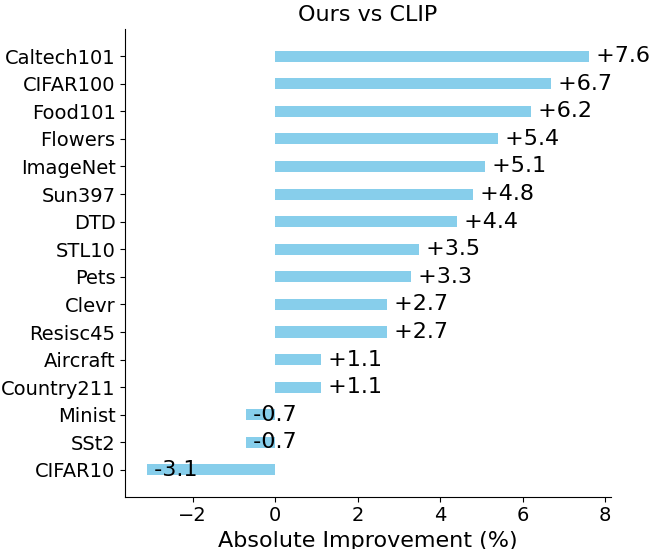}
    \end{minipage}\hfill 
    \begin{minipage}[b]{0.5\linewidth}
        \centering 
        \includegraphics[width=\textwidth]{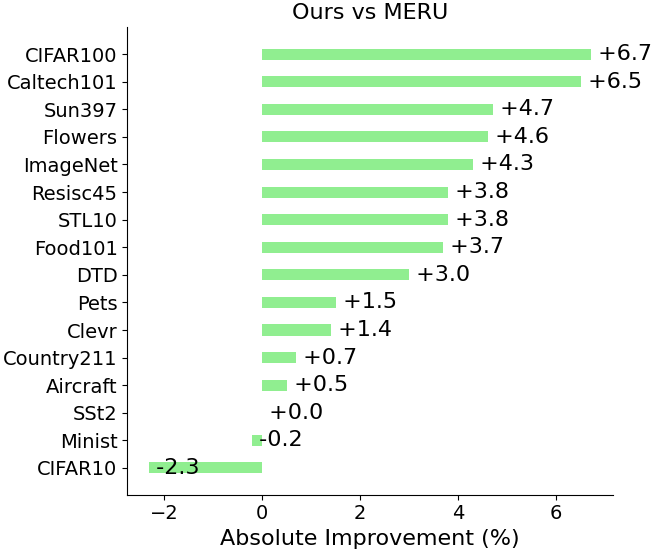}
    \end{minipage}
    \caption{We present the absolute improvements of HMID-Net over CLIP (left) and MERU (right) in zero-shot image classification. HMID-Net achieves +7.6\% improvement over CLIP on the Caltech101 dataset and +6.7\% improvement over MERU on the CIFAR100 dataset.}
    \label{compare baselines}
\end{figure*}

We assess HMID-Net across 16 image classification benchmarks.
Table~\ref{zero-classification} represents the zero-shot image classification performance of HMID-Net. 
We report the absolute improvements of our method over the baseline CLIP and MERU, with the backbone being ViT-L/16.
The left image in Fig.~\ref{compare baselines} compares our method with CLIP, while the right image compares it with MERU.
It significantly outperforms our baselines, CLIP and MERU, on 13 out of the 16 datasets. 
Specifically, HMID-Net achieves +7.6\% improvement over CLIP on the Caltech101 dataset and +6.7\% improvement over MERU on the CIFAR100 dataset.
Additionally, HMID-Net achieves an accuracy of 28.6\% on the ImageNet dataset, surpassing CLIP and MERU by 5.1\% and 4.3\%, respectively.
On the MNIST and SST2 datasets, HMID-Net achieves performance on par with that of CLIP and MERU.
As noted by \cite{hyper-retrieval1}, HMID-Net exhibits relatively suboptimal performance on datasets with fewer covered concepts, such as SST2, which is derived from movie reviews.
Pretraining on larger datasets may improve performance. 
Overall, HMID-Net is highly competitive compared to Euclidean space-based methods.

\begin{table*}[t]
\centering
\renewcommand{\arraystretch}{0.8} 
\begin{tabular}{lccccccccc}
\toprule
& &\multicolumn{4}{c}{\textit{text $\rightarrow$ image}} & \multicolumn{4}{c}{\textit{image $\rightarrow$ text}} \\
\cmidrule(lr){3-6} \cmidrule(lr){7-10}
& &\multicolumn{2}{c}{COCO} & \multicolumn{2}{c}{Flickr} & \multicolumn{2}{c}{COCO} & \multicolumn{2}{c}{Flickr} \\
& &R5 & R10 & R5 & R10 & R5 & R10 & R5 & R10 \\
\midrule
 \multirow{3}{*}{ViT S/16} 
& CLIP & 19.0 & 26.6 & 21.3 & 29.4 & 28.6 & 38.0 & 28.9 & 37.9 \\
& MERU & 19.3 & 27.6 & 22.1 & 30.1 & 28.8 & 37.5 & 26.6 & 35.8 \\
& HMID-Net & \textbf{23.2} & \textbf{31.7} & \textbf{27.7} & \textbf{36.6} & \textbf{33.2} & \textbf{42.9} & \textbf{35.6} & \textbf{45.2} \\
\midrule
\multirow{3}{*}{ViT B/16} 
& CLIP & 20.0 & 26.2 & 22.4 & 30.6 & 28.1 & 38.5 & 32.7 & 41.9 \\
& MERU & 19.5 & 27.8 & 23.1 & 31.1 & 28.6 & 38.6 & 28.2 & 38.2 \\
& HMID-Net & \textbf{24.5} & \textbf{34.0} & \textbf{29.8} & \textbf{38.8} & \textbf{35.6} & \textbf{46.0} & \textbf{37.6} & \textbf{49.3} \\
\midrule
\multirow{3}{*}{ViT L/16} 
& CLIP & 21.2 & 29.9 & 24.2 & 33.5 & 30.9 & 40.8 & 31.3 & 39.2 \\
& MERU & 22.2 & 31.2 & 26.2 & 35.5 & 32.2 & 42.0 & 32.8 & 40.8 \\
& HMID-Net & \cellcolor{mycolor}\textbf{26.9} & \cellcolor{mycolor}\textbf{36.2} & \cellcolor{mycolor}\textbf{33.0} & \cellcolor{mycolor}\textbf{42.1} & \cellcolor{mycolor}\textbf{37.8} & \cellcolor{mycolor}\textbf{48.3} & \cellcolor{mycolor}\textbf{39.6} & \cellcolor{mycolor}\textbf{50.7} \\
\bottomrule
\end{tabular}
\caption{\textbf{Zero-shot image and text retrieval.} HMID-Net achieves the best performance across all retrieval tasks. The best performance in each column is highlighted with \textcolor{mycolor}{orange}.}
\label{retrievaltable}
\end{table*}

\begin{figure*}[tbp!]
    \begin{minipage}[b]{0.49\linewidth}
        \centering 
        \includegraphics[width=\textwidth]{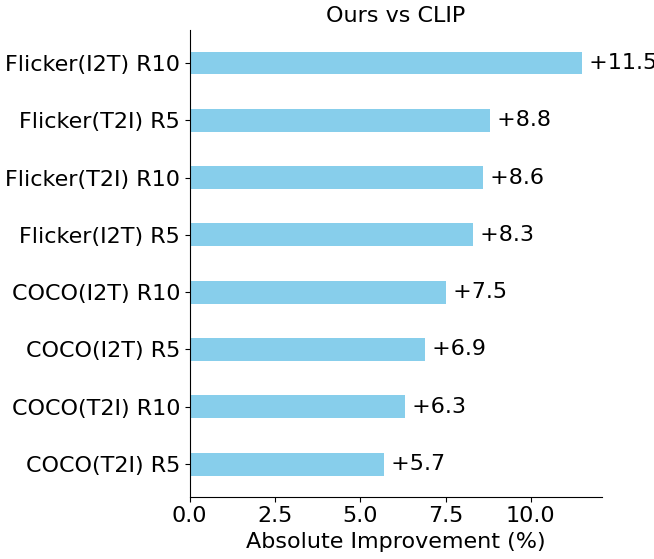}
    \end{minipage}
    \hfill 
    \begin{minipage}[b]{0.5\linewidth}
        \centering 
        \includegraphics[width=\textwidth]{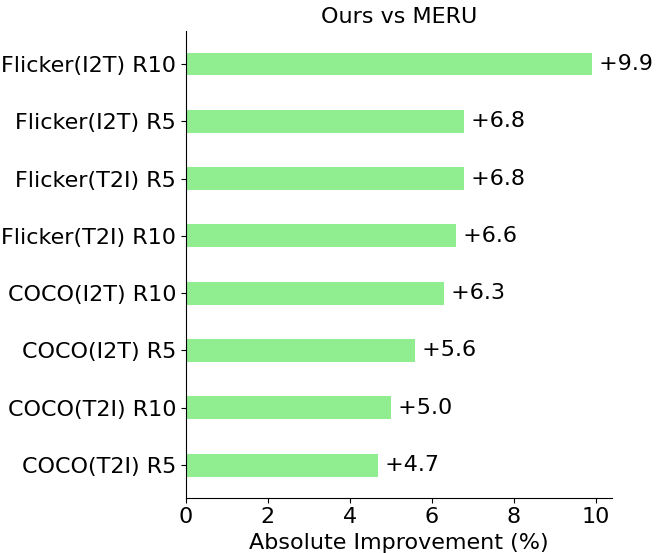}
    \end{minipage}
    \caption{We present the absolute improvements of HMID-Net over CLIP (left) and MERU (right) in zero-shot image and text retrieval. HMID-Net achieves +11.5\% improvement in Flickr (I2T) R\mbox{@}10 over CLIP and +9.9\% improvement in Flickr (I2T) R\mbox{@}10 over MERU.}
    \label{compare retrieval baselines}
\end{figure*}

\begin{table*}[htb]
    \centering 
 
    \subfloat[\textbf{Image masking}      \label{mask}]{\centering
    \begin{minipage}[t]{0.49\linewidth}
      \centering
       \resizebox{\textwidth}{!}{
       \large

      \begin{tabular}{ccccc}
        \toprule
         & \multicolumn{2}{c}{COCO} & \multirow{2}{*}{ImageNet} & \\
        \cmidrule(l){2-3}
         & \textit{text $\to$ image} & \textit{image $\to$ text} & \\
        \midrule
        0\%    & 22.2                     & 32.2                     & 24.3     \\
        25\%   & 22.5                     & 32.4                     & 24.0     \\
        50\%   & 23.5                     & 33.2                     & 26.0     \\
        75\%   & 21.5                     & 30.7                     & 23.0     \\
        \bottomrule
        \end{tabular}
        }
    \end{minipage}%
    }
    \subfloat[\textbf{Unmasked tuning}      \label{tuning}]{\centering
    \begin{minipage}[t]{0.49\linewidth}
      \centering
      \resizebox{\textwidth}{!}{\scriptsize
      \setlength{\heavyrulewidth}{0.65pt} 
      \setlength{\lightrulewidth}{0.45pt} 
      \setlength{\cmidrulewidth}{0.65pt}  
      \begin{tabular}{ccccc}
        \toprule
          & ImageNet & COCO(T2I) &  FLOPs \\
        \midrule
        HMID-Net   & 28.6 & 26.9  &  0.49$\times$ \\
        + tuning   &  28.8 & 27.0  & 1.00$\times$ \\
        \bottomrule
        \end{tabular}}
    \end{minipage}
    }

    \vspace{1cm}

    \begin{subtable}{\textwidth}
        \centering
        \resizebox{\textwidth}{!}{
            \begin{tabular}{@{}cccccccc@{}}
                \toprule
                \multicolumn{2}{c}{Masking ratio} & \multicolumn{3}{c}{Training objective} & \multicolumn{2}{c}{COCO} & \multirow{2}{*}{ImageNet} \\ 
                \cmidrule(r){1-2} \cmidrule(r){3-5} \cmidrule(r){6-7}
                0\% & 50\% & \begin{tabular}[c]{@{}c@{}}contrastive\\ loss\end{tabular} & \begin{tabular}[c]{@{}c@{}}entailment\\ loss\end{tabular} & \begin{tabular}[c]{@{}c@{}}distillation\\ loss\end{tabular} & \textit{text $\to$ image} & \textit{image $\to$ text} & \\ 
                \midrule
                \ding{51} & \ding{55} & \ding{51} & \ding{51} & \ding{55} & 22.2 & 32.2 & 24.3 \\
                \ding{55} & \ding{51} & \ding{51} & \ding{51} & \ding{55} & 23.5 & 33.2 & 26.0 \\
                \ding{55} & \ding{51} & \ding{51} & \ding{55} & \ding{51} & 27.1 & 38.4 & 23.0 \\
                \ding{55} & \ding{51} & \ding{51} & \ding{51} & \ding{51} & 26.9 & 37.8 & 28.6 \\ 
                \bottomrule
            \end{tabular}}
        \subcaption{\textbf{Loss function}} \label{Loss function}
    \end{subtable}
    \caption{\textbf{Ablation experiments.} The backbone is ViT-L/16. Unless specified otherwise, the default configuration is: image masking is 50\% and no unmasked tuning.}
    \label{Ablations Experiment}
\end{table*}

\subsection{Image and text retrieval}
\label{retrieval}
CLIP-style contrastive models pull image-text pairs with high similarity closer together during training, while pushing those with dissimilarity farther apart.
This approach is highly beneficial for retrieval tasks.
We evaluate HMID-Net on two benchmarks: COCO and Flickr30K.
We report the recall\mbox{@}\{5,10\} performance in Table~\ref{retrievaltable}.
HMID-Net, trained with ViT of varying parameter sizes, achieves the best performance in both T2I and I2T retrieval tasks, significantly outperforming the baseline methods CLIP and MERU.
Fig.~\ref{compare retrieval baselines} illustrates the comparison of HMID-Net with CLIP (left) and MERU (right) in zero-shot image and text retrieval, with the backbone being ViT-L/16.
HMID-Net achieves +11.5\% improvement in Flickr (I2T) R\mbox{@}10 over CLIP and +9.9\% improvement in Flickr (I2T) R\mbox{@}10 over MERU.
This demonstrates that the geometric properties of hyperbolic space facilitate the learning of more robust and effective representations for retrieval tasks.

\subsection{Ablations experiments}
\label{ablations}
We perform an ablation study on our HMID-Net model to evaluate the impact of the designed modules.
Our ablation experiments are conducted on ViT-L/16, utilizing the ImageNet dataset and the COCO dataset for zero-shot evaluation.
We report the zero-shot COCO recall@5 for retrieval tasks and the zero-shot top-1 accuracy for classification tasks, as presented in Table~\ref{Ablations Experiment}.

\textbf{Masking ratio.} 
We first conducted a study on image masking ratios based solely on MERU, as shown in Table~\ref{mask}.
The 0\% masking ratio refers to our baseline MERU.
A 50\% masking ratio yields optimal performance, with a 1.7\% improvement on ImageNet and a 1.3\% improvement on COCO for the T2I retrieval task. 
Compared to BERT's 15\% masking ratio, images have significant pixel redundancy, enabling a higher masking ratio. 
However, a 75\% masking ratio results in a performance decline due to the loss of critical information, which hinders contrastive learning.
This effect is similar to the findings in FLIP \cite{flip}. 
Unless specified otherwise, we use a default masking ratio of 50\%.

\textbf{Unmasked tuning.}
During pretraining, we use masked images, while during inference, complete unmasked images are input. 
We examine the gap between pretraining and inference.
Table~\ref{tuning} shows results from an additional 0.5 epoch of fine-tuning on unmasked images during pretraining.
Fine-tuning yields a 0.2\% and 0.1\% performance increase on ImageNet and COCO T2I, respectively, narrowing the gap.
We measure the FLOPs of the model's visual component, finding that fine-tuning with unmasked images doubles the FLOPs compared to masked image training. 
Although fine-tuning offers a slight performance gain, it significantly increases computational cost. 
Thus, we opt for masked images to balance performance and computational efficiency.

\textbf{Loss Function.} 
We further investigated the impact of contrastive loss, entailment loss, and distillation loss.
Table~\ref{Loss function} presents the experimental details.
The first row represents our baseline MERU, which uses only contrastive and entailment losses.
With a 50\% masking ratio, significant improvements are observed in both COCO retrieval and ImageNet classification. 
Surprisingly, when the loss function includes only contrastive and distillation losses, ImageNet classification performance drops by 1.3\% compared to MERU, while COCO text-to-image retrieval improves by 4.9\%.  
Our model, HMID-Net (last row), incorporates contrastive, entailment, and distillation losses.
Although there is a slight decrease in COCO retrieval performance compared to the scenario without entailment loss, a 5.6\% improvement is achieved in ImageNet classification.
HMID-Net outperforms MERU by 4.3\% on ImageNet classification and by 4.7\% on COCO T2I retrieval, demonstrating the effectiveness of image masking and knowledge distillation in hyperbolic space.

\subsection{Qualitative analysis}
\label{Qualitative analysis}
This section presents a qualitative analysis of the visual-semantic hierarchy. 
General objects are closer to the [Root], while specific objects are near the boundary \cite{hyper-retrieval2}.
The distance from the origin indicates uncertainty, useful for retrieval tasks. 
In this hierarchy, text is closer to the origin and images nearer the boundary.
For example, in Fig.~\ref{fig1}(a), the "Cheerful smile" is near the [ROOT], while the image is positioned closer to the boundary.

\begin{figure*}[tbp]
\centering
\includegraphics[width=1\textwidth]{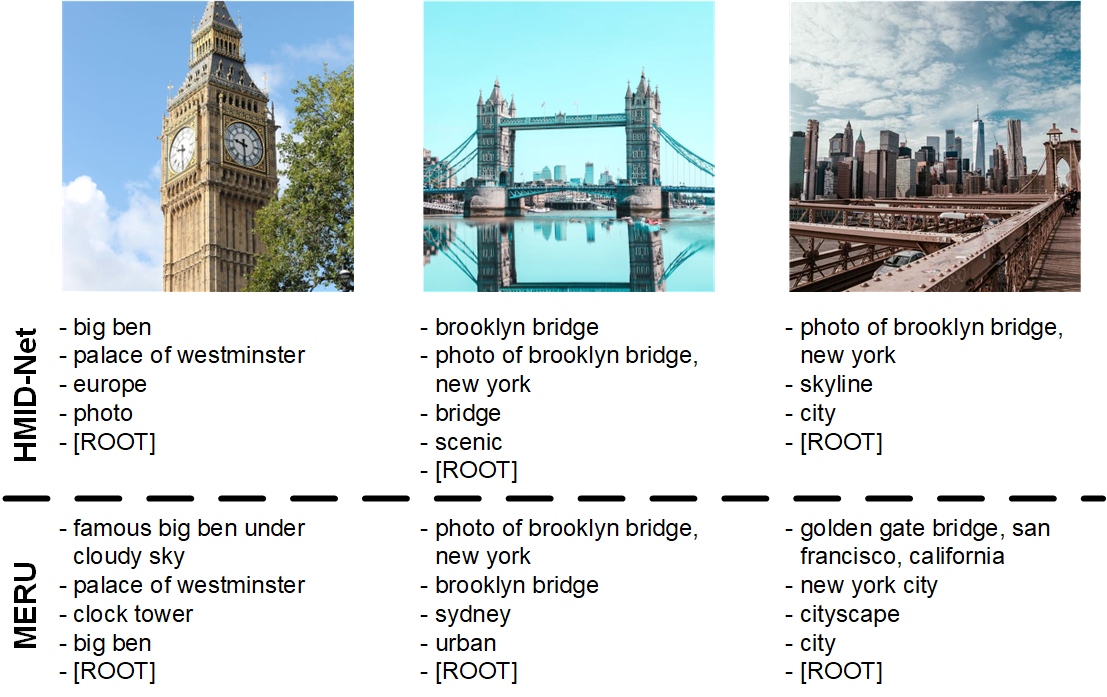}
\caption{\textbf{An illustration of visual-semantic hierarchy.} We performed text retrieval at each interpolation step, selecting the description with the highest Lorentzian inner product. As the embedding approaches the [ROOT], the descriptions become more general.}
\label{qualitative}
\end{figure*}

Our experiment closely follows MERU \cite{hyper-retrieval1}. 
A subset of images is randomly selected from Pixels, with textual descriptions retrieved from a curated set of 750 captions on pexels.com.
We interpolate 50 steps along the geodesic between the image embedding and the [ROOT], selecting the textual description with the highest Lorentzian inner product at each step.
Duplicates are removed, and the top five descriptions are retained.
Fig.~\ref{qualitative} shows that both HMID-Net and MERU effectively capture the visual-semantic hierarchy, with descriptions becoming more generic closer to the [ROOT].

\section{Discussion and Conclusion}
\label{conclusion}

In this paper, we propose the Hyperbolic Masked Image and Distillation Network (HMID-Net), which integrates Masked Image Modeling (MIM) and knowledge distillation in hyperbolic space to more efficiently learn the visual-semantic hierarchy. 
Our method introduces knowledge distillation to hyperbolic space, achieving training efficiency on par with Euclidean space. Experimental results demonstrate that HMID-Net enhances real-world understanding and outperforms baseline models MERU and CLIP across a range of tasks, highlighting the effectiveness of MIM and knowledge distillation in hyperbolic space.
However, our method still have several limitations.
First, HMID-Net exhibits slower convergence during training, which we attribute to the difficulty of jointly optimizing multiple loss functions.
The presence of multiple objectives increases the complexity of the optimization landscape, making it challenging for the model to locate a single optimal solution that balances all tasks effectively.
Second, the performance of HMID-Net on datasets such as SST-2 remains suboptimal. 
We believe this is primarily due to the limited coverage of SST-2-like samples in the current training dataset, which hinders the model's ability to generalize to sentiment classification tasks that require sensitivity to fine-grained emotional and linguistic cues.

In future work, to address the slow convergence issue, we plan to refine the knowledge distillation loss function, aiming to reduce conflicts among multiple objectives and facilitate more stable and efficient optimization.
By better aligning the learning signals from different supervision sources, we expect to accelerate convergence during training.
To improve the generalization performance on datasets such as SST-2, we intend to expand and diversify the training data by incorporating large-scale, sentiment-rich datasets. 
This enhancement will allow the model to learn from a broader distribution of linguistic patterns and emotional expressions, thereby improving its adaptability and performance across a wider range of downstream tasks.

\newpage
\newpage

\bibliography{aaai2026}

\begin{thebibliography}{57}
\providecommand{\natexlab}[1]{#1}

\bibitem[{Atigh et~al.(2022)Atigh, Schoep, Acar, Van~Noord, and Mettes}]{hyper-segmentation}
Atigh, M.~G.; Schoep, J.; Acar, E.; Van~Noord, N.; and Mettes, P. 2022.
\newblock Hyperbolic image segmentation.
\newblock In \emph{Proceedings of the IEEE/CVF conference on computer vision and pattern recognition}, 4453--4462.

\bibitem[{Baldrati et~al.(2023)Baldrati, Bertini, Uricchio, and Del~Bimbo}]{ref7}
Baldrati, A.; Bertini, M.; Uricchio, T.; and Del~Bimbo, A. 2023.
\newblock Composed image retrieval using contrastive learning and task-oriented clip-based features.
\newblock \emph{ACM Transactions on Multimedia Computing, Communications and Applications}, 20(3): 1--24.

\bibitem[{Bao et~al.(2022)Bao, Dong, Piao, and Wei}]{beit}
Bao, H.; Dong, L.; Piao, S.; and Wei, F. 2022.
\newblock {BEiT}: {BERT} Pre-Training of Image Transformers.
\newblock In \emph{International Conference on Learning Representations}.

\bibitem[{Bossard, Guillaumin, and Van~Gool(2014)}]{food}
Bossard, L.; Guillaumin, M.; and Van~Gool, L. 2014.
\newblock Food-101--mining discriminative components with random forests.
\newblock In \emph{Computer vision--ECCV 2014: 13th European conference, zurich, Switzerland, September 6-12, 2014, proceedings, part VI 13}, 446--461. Springer.

\bibitem[{Chen et~al.(2015)Chen, Fang, Lin, Vedantam, Gupta, Doll{\'a}r, and Zitnick}]{coco}
Chen, X.; Fang, H.; Lin, T.-Y.; Vedantam, R.; Gupta, S.; Doll{\'a}r, P.; and Zitnick, C.~L. 2015.
\newblock Microsoft coco captions: Data collection and evaluation server.
\newblock \emph{arXiv preprint arXiv:1504.00325}.

\bibitem[{Cheng, Han, and Lu(2017)}]{Resisc45}
Cheng, G.; Han, J.; and Lu, X. 2017.
\newblock Remote sensing image scene classification: Benchmark and state of the art.
\newblock \emph{Proceedings of the IEEE}, 105(10): 1865--1883.

\bibitem[{Cherti et~al.(2023)Cherti, Beaumont, Wightman, Wortsman, Ilharco, Gordon, Schuhmann, Schmidt, and Jitsev}]{open-clip}
Cherti, M.; Beaumont, R.; Wightman, R.; Wortsman, M.; Ilharco, G.; Gordon, C.; Schuhmann, C.; Schmidt, L.; and Jitsev, J. 2023.
\newblock Reproducible scaling laws for contrastive language-image learning.
\newblock In \emph{Proceedings of the IEEE/CVF Conference on Computer Vision and Pattern Recognition}, 2818--2829.

\bibitem[{Cimpoi et~al.(2014)Cimpoi, Maji, Kokkinos, Mohamed, and Vedaldi}]{dtd}
Cimpoi, M.; Maji, S.; Kokkinos, I.; Mohamed, S.; and Vedaldi, A. 2014.
\newblock Describing textures in the wild.
\newblock In \emph{Proceedings of the IEEE conference on computer vision and pattern recognition}, 3606--3613.

\bibitem[{Coates, Ng, and Lee(2011)}]{stl10}
Coates, A.; Ng, A.; and Lee, H. 2011.
\newblock An analysis of single-layer networks in unsupervised feature learning.
\newblock In \emph{Proceedings of the fourteenth international conference on artificial intelligence and statistics}, 215--223. JMLR Workshop and Conference Proceedings.

\bibitem[{Deng et~al.(2009)Deng, Dong, Socher, Li, Li, and Fei-Fei}]{imagenet}
Deng, J.; Dong, W.; Socher, R.; Li, L.-J.; Li, K.; and Fei-Fei, L. 2009.
\newblock Imagenet: A large-scale hierarchical image database.
\newblock In \emph{2009 IEEE conference on computer vision and pattern recognition}, 248--255. Ieee.

\bibitem[{Desai et~al.(2023)Desai, Nickel, Rajpurohit, Johnson, and Vedantam}]{hyper-retrieval1}
Desai, K.; Nickel, M.; Rajpurohit, T.; Johnson, J.; and Vedantam, S.~R. 2023.
\newblock Hyperbolic image-text representations.
\newblock In \emph{International Conference on Machine Learning}, 7694--7731. PMLR.

\bibitem[{Dhall et~al.(2020)Dhall, Makarova, Ganea, Pavllo, Greeff, and Krause}]{hyper-classfication3}
Dhall, A.; Makarova, A.; Ganea, O.; Pavllo, D.; Greeff, M.; and Krause, A. 2020.
\newblock Hierarchical image classification using entailment cone embeddings.
\newblock In \emph{Proceedings of the IEEE/CVF conference on computer vision and pattern recognition workshops}, 836--837.

\bibitem[{Dosovitskiy et~al.(2021)Dosovitskiy, Beyer, Kolesnikov, Weissenborn, Zhai, Unterthiner, Dehghani, Minderer, Heigold, Gelly, Uszkoreit, and Houlsby}]{vit}
Dosovitskiy, A.; Beyer, L.; Kolesnikov, A.; Weissenborn, D.; Zhai, X.; Unterthiner, T.; Dehghani, M.; Minderer, M.; Heigold, G.; Gelly, S.; Uszkoreit, J.; and Houlsby, N. 2021.
\newblock An Image is Worth 16x16 Words: Transformers for Image Recognition at Scale.
\newblock \emph{ICLR}.

\bibitem[{Fei-Fei, Fergus, and Perona(2004)}]{Caltech101}
Fei-Fei, L.; Fergus, R.; and Perona, P. 2004.
\newblock Learning generative visual models from few training examples: An incremental bayesian approach tested on 101 object categories.
\newblock In \emph{2004 conference on computer vision and pattern recognition workshop}, 178--178. IEEE.

\bibitem[{Feng et~al.(2025)Feng, Lin, Tang, Wang, Zheng, He, Pang, Yang, Chen, and Wei}]{hyperbolic-pointcloud}
Feng, Y.-Z.; Lin, S.-H.~J.; Tang, X.; Wang, M.-Y.; Zheng, J.-Z.; He, Z.-Y.; Pang, Z.-Y.; Yang, J.; Chen, M.-S.; and Wei, X. 2025.
\newblock Hyperbolic prototype rectification for few-shot 3D point cloud classification.
\newblock \emph{Pattern Recognition}, 158: 111042.

\bibitem[{Ganea, B{\'e}cigneul, and Hofmann(2018{\natexlab{a}})}]{ref8}
Ganea, O.; B{\'e}cigneul, G.; and Hofmann, T. 2018{\natexlab{a}}.
\newblock Hyperbolic entailment cones for learning hierarchical embeddings.
\newblock In \emph{International conference on machine learning}, 1646--1655. PMLR.

\bibitem[{Ganea, B{\'e}cigneul, and Hofmann(2018{\natexlab{b}})}]{hyper3}
Ganea, O.; B{\'e}cigneul, G.; and Hofmann, T. 2018{\natexlab{b}}.
\newblock Hyperbolic neural networks.
\newblock \emph{Advances in neural information processing systems}, 31.

\bibitem[{Gao et~al.(2024)Gao, Geng, Zhang, Ma, Fang, Zhang, Li, and Qiao}]{clip-adapter}
Gao, P.; Geng, S.; Zhang, R.; Ma, T.; Fang, R.; Zhang, Y.; Li, H.; and Qiao, Y. 2024.
\newblock Clip-adapter: Better vision-language models with feature adapters.
\newblock \emph{International Journal of Computer Vision}, 132(2): 581--595.

\bibitem[{Gu et~al.(2022)Gu, Lin, Kuo, and Cui}]{ref3}
Gu, X.; Lin, T.-Y.; Kuo, W.; and Cui, Y. 2022.
\newblock Open-vocabulary Object Detection via Vision and Language Knowledge Distillation.
\newblock In \emph{International Conference on Learning Representations}.

\bibitem[{Guo et~al.(2023)Guo, Zhang, Qiu, Ma, Miao, He, and Cui}]{calip}
Guo, Z.; Zhang, R.; Qiu, L.; Ma, X.; Miao, X.; He, X.; and Cui, B. 2023.
\newblock Calip: Zero-shot enhancement of clip with parameter-free attention.
\newblock In \emph{Proceedings of the AAAI Conference on Artificial Intelligence}, volume~37, 746--754.

\bibitem[{He et~al.(2022)He, Chen, Xie, Li, Doll{\'a}r, and Girshick}]{mae}
He, K.; Chen, X.; Xie, S.; Li, Y.; Doll{\'a}r, P.; and Girshick, R. 2022.
\newblock Masked autoencoders are scalable vision learners.
\newblock In \emph{Proceedings of the IEEE/CVF conference on computer vision and pattern recognition}, 16000--16009.

\bibitem[{He et~al.(2016)He, Zhang, Ren, and Sun}]{resnet}
He, K.; Zhang, X.; Ren, S.; and Sun, J. 2016.
\newblock Deep residual learning for image recognition.
\newblock In \emph{Proceedings of the IEEE conference on computer vision and pattern recognition}, 770--778.

\bibitem[{Jia et~al.(2021)Jia, Yang, Xia, Chen, Parekh, Pham, Le, Sung, Li, and Duerig}]{ALIGN}
Jia, C.; Yang, Y.; Xia, Y.; Chen, Y.-T.; Parekh, Z.; Pham, H.; Le, Q.; Sung, Y.-H.; Li, Z.; and Duerig, T. 2021.
\newblock Scaling up visual and vision-language representation learning with noisy text supervision.
\newblock In \emph{International conference on machine learning}, 4904--4916. PMLR.

\bibitem[{Jiang et~al.(2025)Jiang, Zhang, Huang, Ge, Ni, Song, and Huang}]{adapter-retrieval}
Jiang, H.; Zhang, J.; Huang, R.; Ge, C.; Ni, Z.; Song, S.; and Huang, G. 2025.
\newblock Cross-modal adapter for vision--language retrieval.
\newblock \emph{Pattern Recognition}, 159: 111144.

\bibitem[{Johnson et~al.(2017)Johnson, Hariharan, Van Der~Maaten, Fei-Fei, Lawrence~Zitnick, and Girshick}]{clevr}
Johnson, J.; Hariharan, B.; Van Der~Maaten, L.; Fei-Fei, L.; Lawrence~Zitnick, C.; and Girshick, R. 2017.
\newblock Clevr: A diagnostic dataset for compositional language and elementary visual reasoning.
\newblock In \emph{Proceedings of the IEEE conference on computer vision and pattern recognition}, 2901--2910.

\bibitem[{Khrulkov et~al.(2020)Khrulkov, Mirvakhabova, Ustinova, Oseledets, and Lempitsky}]{hyper-classfication1}
Khrulkov, V.; Mirvakhabova, L.; Ustinova, E.; Oseledets, I.; and Lempitsky, V. 2020.
\newblock Hyperbolic image embeddings.
\newblock In \emph{Proceedings of the IEEE/CVF conference on computer vision and pattern recognition}, 6418--6428.

\bibitem[{Kong et~al.(2024)Kong, Chen, Cai, and Modolo}]{hyper-detection}
Kong, F.; Chen, Y.; Cai, J.; and Modolo, D. 2024.
\newblock Hyperbolic learning with synthetic captions for open-world detection.
\newblock In \emph{Proceedings of the IEEE/CVF Conference on Computer Vision and Pattern Recognition}, 16762--16771.

\bibitem[{Krizhevsky, Hinton et~al.(2009)}]{cifar}
Krizhevsky, A.; Hinton, G.; et~al. 2009.
\newblock Learning multiple layers of features from tiny images.

\bibitem[{Kwon et~al.(2024)Kwon, Jang, Kim, Kim, and Sohn}]{hyper-classfication4}
Kwon, H.; Jang, J.; Kim, J.; Kim, K.; and Sohn, K. 2024.
\newblock Improving Visual Recognition with Hyperbolical Visual Hierarchy Mapping.
\newblock In \emph{Proceedings of the IEEE/CVF Conference on Computer Vision and Pattern Recognition}, 17364--17374.

\bibitem[{LeCun et~al.(1998)LeCun, Bottou, Bengio, and Haffner}]{minist}
LeCun, Y.; Bottou, L.; Bengio, Y.; and Haffner, P. 1998.
\newblock Gradient-based learning applied to document recognition.
\newblock \emph{Proceedings of the IEEE}, 86(11): 2278--2324.

\bibitem[{Li* et~al.(2022)Li*, Zhang*, Zhang*, Yang, Li, Zhong, Wang, Yuan, Zhang, Hwang, Chang, and Gao}]{ref4}
Li*, L.~H.; Zhang*, P.; Zhang*, H.; Yang, J.; Li, C.; Zhong, Y.; Wang, L.; Yuan, L.; Zhang, L.; Hwang, J.-N.; Chang, K.-W.; and Gao, J. 2022.
\newblock Grounded Language-Image Pre-training.
\newblock In \emph{CVPR}.

\bibitem[{Li et~al.(2023)Li, Fan, Hu, Feichtenhofer, and He}]{flip}
Li, Y.; Fan, H.; Hu, R.; Feichtenhofer, C.; and He, K. 2023.
\newblock Scaling language-image pre-training via masking.
\newblock In \emph{Proceedings of the IEEE/CVF Conference on Computer Vision and Pattern Recognition}, 23390--23400.

\bibitem[{Liu et~al.(2020)Liu, Chen, Pan, Ngo, Chua, and Jiang}]{hyper-classfication2}
Liu, S.; Chen, J.; Pan, L.; Ngo, C.-W.; Chua, T.-S.; and Jiang, Y.-G. 2020.
\newblock Hyperbolic visual embedding learning for zero-shot recognition.
\newblock In \emph{Proceedings of the IEEE/CVF conference on computer vision and pattern recognition}, 9273--9281.

\bibitem[{Loshchilov and Hutter(2019)}]{adaw}
Loshchilov, I.; and Hutter, F. 2019.
\newblock Decoupled Weight Decay Regularization.
\newblock In \emph{7th International Conference on Learning Representations, {ICLR} 2019, New Orleans, LA, USA, May 6-9, 2019}. OpenReview.net.

\bibitem[{Luo et~al.(2022)Luo, Ji, Zhong, Chen, Lei, Duan, and Li}]{ref5}
Luo, H.; Ji, L.; Zhong, M.; Chen, Y.; Lei, W.; Duan, N.; and Li, T. 2022.
\newblock Clip4clip: An empirical study of clip for end to end video clip retrieval and captioning.
\newblock \emph{Neurocomputing}, 508: 293--304.

\bibitem[{Maji et~al.(2013)Maji, Rahtu, Kannala, Blaschko, and Vedaldi}]{aircraft}
Maji, S.; Rahtu, E.; Kannala, J.; Blaschko, M.; and Vedaldi, A. 2013.
\newblock Fine-grained visual classification of aircraft.
\newblock \emph{arXiv preprint arXiv:1306.5151}.

\bibitem[{Nickel and Kiela(2017)}]{hyper1}
Nickel, M.; and Kiela, D. 2017.
\newblock Poincar{\'e} embeddings for learning hierarchical representations.
\newblock \emph{Advances in neural information processing systems}, 30.

\bibitem[{Nilsback and Zisserman(2008)}]{Flowers}
Nilsback, M.-E.; and Zisserman, A. 2008.
\newblock Automated flower classification over a large number of classes.
\newblock In \emph{2008 Sixth Indian conference on computer vision, graphics \& image processing}, 722--729. IEEE.

\bibitem[{Parkhi et~al.(2012)Parkhi, Vedaldi, Zisserman, and Jawahar}]{pets}
Parkhi, O.~M.; Vedaldi, A.; Zisserman, A.; and Jawahar, C. 2012.
\newblock Cats and dogs.
\newblock In \emph{2012 IEEE conference on computer vision and pattern recognition}, 3498--3505. IEEE.

\bibitem[{Peng et~al.(2021)Peng, Varanka, Mostafa, Shi, and Zhao}]{hyper2}
Peng, W.; Varanka, T.; Mostafa, A.; Shi, H.; and Zhao, G. 2021.
\newblock Hyperbolic deep neural networks: A survey.
\newblock \emph{IEEE Transactions on pattern analysis and machine intelligence}, 44(12): 10023--10044.

\bibitem[{Radford et~al.(2021)Radford, Kim, Hallacy, Ramesh, Goh, Agarwal, Sastry, Askell, Mishkin, Clark et~al.}]{clip}
Radford, A.; Kim, J.~W.; Hallacy, C.; Ramesh, A.; Goh, G.; Agarwal, S.; Sastry, G.; Askell, A.; Mishkin, P.; Clark, J.; et~al. 2021.
\newblock Learning transferable visual models from natural language supervision.
\newblock In \emph{International conference on machine learning}, 8748--8763. PMLR.

\bibitem[{Ramasinghe et~al.(2024)Ramasinghe, Shevchenko, Avraham, and Thalaiyasingam}]{hyper-retrieval2}
Ramasinghe, S.; Shevchenko, V.; Avraham, G.; and Thalaiyasingam, A. 2024.
\newblock Accept the modality gap: An exploration in the hyperbolic space.
\newblock In \emph{Proceedings of the IEEE/CVF Conference on Computer Vision and Pattern Recognition}, 27263--27272.

\bibitem[{Vaswani et~al.(2017)Vaswani, Shazeer, Parmar, Uszkoreit, Jones, Gomez, Kaiser, and Polosukhin}]{transformer}
Vaswani, A.; Shazeer, N.; Parmar, N.; Uszkoreit, J.; Jones, L.; Gomez, A.~N.; Kaiser, {\L}.; and Polosukhin, I. 2017.
\newblock Attention is all you need.
\newblock \emph{Advances in neural information processing systems}, 30.

\bibitem[{Vendrov et~al.(2016)Vendrov, Kiros, Fidler, and Urtasun}]{ref1}
Vendrov, I.; Kiros, R.; Fidler, S.; and Urtasun, R. 2016.
\newblock {Order-embeddings of images and language}.
\newblock In \emph{International Conference on Learning Representations}.

\bibitem[{Wei et~al.(2022)Wei, Fan, Xie, Wu, Yuille, and Feichtenhofer}]{hog}
Wei, C.; Fan, H.; Xie, S.; Wu, C.-Y.; Yuille, A.; and Feichtenhofer, C. 2022.
\newblock Masked feature prediction for self-supervised visual pre-training.
\newblock In \emph{Proceedings of the IEEE/CVF Conference on Computer Vision and Pattern Recognition}, 14668--14678.

\bibitem[{Wu et~al.(2023)Wu, Peng, Zhou, Xiao, Liu, Yuan, Xuan, Valenzuela, Chen, Wang et~al.}]{tinyclip}
Wu, K.; Peng, H.; Zhou, Z.; Xiao, B.; Liu, M.; Yuan, L.; Xuan, H.; Valenzuela, M.; Chen, X.~S.; Wang, X.; et~al. 2023.
\newblock Tinyclip: Clip distillation via affinity mimicking and weight inheritance.
\newblock In \emph{Proceedings of the IEEE/CVF International Conference on Computer Vision}, 21970--21980.

\bibitem[{Wu et~al.(2025)Wu, Zhang, Zhang, Zhao, Liang, and Yang}]{textkd}
Wu, L.; Zhang, S.; Zhang, C.; Zhao, Z.; Liang, J.; and Yang, W. 2025.
\newblock Enhancing knowledge distillation for semantic segmentation through text-assisted modular plugins.
\newblock \emph{Pattern Recognition}, 161: 111329.

\bibitem[{Xiao et~al.(2010)Xiao, Hays, Ehinger, Oliva, and Torralba}]{sun}
Xiao, J.; Hays, J.; Ehinger, K.~A.; Oliva, A.; and Torralba, A. 2010.
\newblock Sun database: Large-scale scene recognition from abbey to zoo.
\newblock In \emph{2010 IEEE computer society conference on computer vision and pattern recognition}, 3485--3492. IEEE.

\bibitem[{Xie et~al.(2022)Xie, Zhang, Cao, Lin, Bao, Yao, Dai, and Hu}]{simmim}
Xie, Z.; Zhang, Z.; Cao, Y.; Lin, Y.; Bao, J.; Yao, Z.; Dai, Q.; and Hu, H. 2022.
\newblock Simmim: A simple framework for masked image modeling.
\newblock In \emph{Proceedings of the IEEE/CVF conference on computer vision and pattern recognition}, 9653--9663.

\bibitem[{Yang et~al.(2024)Yang, An, Huang, Bi, Yu, Yang, Diao, and Xu}]{clipkd}
Yang, C.; An, Z.; Huang, L.; Bi, J.; Yu, X.; Yang, H.; Diao, B.; and Xu, Y. 2024.
\newblock CLIP-KD: An Empirical Study of CLIP Model Distillation.
\newblock In \emph{Proceedings of the IEEE/CVF Conference on Computer Vision and Pattern Recognition}, 15952--15962.

\bibitem[{Young et~al.(2014)Young, Lai, Hodosh, and Hockenmaier}]{Flickr30k}
Young, P.; Lai, A.; Hodosh, M.; and Hockenmaier, J. 2014.
\newblock From image descriptions to visual denotations: New similarity metrics for semantic inference over event descriptions.
\newblock \emph{Transactions of the association for computational linguistics}, 2: 67--78.

\bibitem[{Zhang et~al.(2019)Zhang, Song, Gao, Chen, Bao, and Ma}]{self-distill}
Zhang, L.; Song, J.; Gao, A.; Chen, J.; Bao, C.; and Ma, K. 2019.
\newblock Be your own teacher: Improve the performance of convolutional neural networks via self distillation.
\newblock In \emph{Proceedings of the IEEE/CVF international conference on computer vision}, 3713--3722.

\bibitem[{Zhang, Wang, and Wang(2022)}]{ref11}
Zhang, Q.; Wang, Y.; and Wang, Y. 2022.
\newblock How mask matters: Towards theoretical understandings of masked autoencoders.
\newblock \emph{Advances in Neural Information Processing Systems}, 35: 27127--27139.

\bibitem[{Zhang et~al.(2022)Zhang, Zhang, Fang, Gao, Li, Dai, Qiao, and Li}]{Tip-Adapter}
Zhang, R.; Zhang, W.; Fang, R.; Gao, P.; Li, K.; Dai, J.; Qiao, Y.; and Li, H. 2022.
\newblock Tip-adapter: Training-free adaption of clip for few-shot classification.
\newblock In \emph{European conference on computer vision}, 493--510. Springer.

\bibitem[{Zhao et~al.(2022)Zhao, Zhu, Wang, and Yang}]{ref6}
Zhao, S.; Zhu, L.; Wang, X.; and Yang, Y. 2022.
\newblock Centerclip: Token clustering for efficient text-video retrieval.
\newblock In \emph{Proceedings of the 45th International ACM SIGIR Conference on Research and Development in Information Retrieval}, 970--981.

\bibitem[{Zhou et~al.(2022{\natexlab{a}})Zhou, Yang, Loy, and Liu}]{cocoop}
Zhou, K.; Yang, J.; Loy, C.~C.; and Liu, Z. 2022{\natexlab{a}}.
\newblock Conditional prompt learning for vision-language models.
\newblock In \emph{Proceedings of the IEEE/CVF conference on computer vision and pattern recognition}, 16816--16825.

\bibitem[{Zhou et~al.(2022{\natexlab{b}})Zhou, Yang, Loy, and Liu}]{coop}
Zhou, K.; Yang, J.; Loy, C.~C.; and Liu, Z. 2022{\natexlab{b}}.
\newblock Learning to prompt for vision-language models.
\newblock \emph{International Journal of Computer Vision}, 130(9): 2337--2348.

\end{thebibliography}

\end{document}